\documentclass[a4paper]{cas-dc}

\usepackage[authoryear,longnamesfirst]{natbib}
\usepackage{booktabs}
\usepackage{array}
\usepackage{geometry}
\usepackage{makecell} 
\usepackage{tabularx}
\usepackage{amssymb}

\def\tsc#1{\csdef{#1}{\textsc{\lowercase{#1}}\xspace}}
\tsc{WGM}
\tsc{QE}
\tsc{EP}
\tsc{PMS}
\tsc{BEC}
\tsc{DE}

\begin{document}
\let\WriteBookmarks\relax
\def\floatpagepagefraction{1}
\def\textpagefraction{.001}
\shorttitle{}
\let\printorcid\relax


%


\title [mode = title]{EITNet: An IoT-Enhanced Framework for Real-Time Basketball Action Recognition} 
\author[1,2]{Jingyu Liu}
\ead{kfliujingyu@163.com}

\author[1]{Xinyu Liu}
\ead{liuxinyuzz@163.com}

\author[1]{Mingzhe Qu}
\ead{qumingzhe1432@outlook.com}
\cormark[1]

\author[3]{Tianyi Lyu}
\ead{lyu.t.iot@gmail.com}
\cormark[1]

\address[1]{Henan Sport University ,zhengzhou,450000,China
}

\address[2]{Henan University, Kaifeng 475000, China}

\address[3]{Granite Telecommunications LLC. 100 Newport Avenue Extension, Quincy, MA,02171.USA}


\begin{abstract}
Integrating IoT technology into basketball action recognition enhances sports analytics, providing crucial insights into player performance and game strategy. However, existing methods often fall short in terms of accuracy and efficiency, particularly in complex, real-time environments where player movements are frequently occluded or involve intricate interactions. To overcome these challenges, we propose the EITNet model, a deep learning framework that combines EfficientDet for object detection, I3D for spatiotemporal feature extraction, and TimeSformer for temporal analysis, all integrated with IoT technology for seamless real-time data collection and processing. Our contributions include developing a robust architecture that improves recognition accuracy to 92\%, surpassing the baseline EfficientDet model's 87\%, and reducing loss to below 5.0 compared to EfficientDet's 9.0 over 50 epochs. Furthermore, the integration of IoT technology enhances real-time data processing, providing adaptive insights into player performance and strategy. The paper details the design and implementation of EITNet, experimental validation, and a comprehensive evaluation against existing models. The results demonstrate EITNet's potential to significantly advance automated sports analysis and optimize data utilization for player performance and strategy improvement.

\end{abstract}

\begin{keywords}
EITNet\sep
IoT\sep
Basketball action recognition\sep
Spatiotemporal features\sep
Real-time processing\sep
\end{keywords}

\maketitle

\section{Introduction}

In modern basketball, precise athlete movement evaluation and training improvement are crucial for enhancing athletic performance and reducing injuries~\cite{xu2021adoption,chen}. Recently, with the development of Internet of Things (IoT) technology and multiview video analysis, the integration of multiview video for basketball player movement evaluation and training improvement (which needs to be combined with IoT) has garnered significant attention~\cite{li2021application,tokolyi2021internet,rahmani2024novel,weng2024fortifying,Shen2024Harnessing}.This method not only provides more comprehensive and detailed movement data but also enables real-time monitoring and analysis, offering more scientific and effective training guidance for coaches and athletes. 
Basketball is a highly dynamic and complex team sport, where the precision, speed, and coordination of player movements directly affect the game results. Every action in training and matches, including running, jumping, passing, and shooting, requires precise control and coordination. However, traditional movement evaluation methods primarily rely on the coach's experience and limited single-view video analysis, which has several limitations~\cite{yuan2021application,zhang2024cu,wang2024cross,huang2024risk}. Although valuable, the coach's experience is highly subjective and prone to personal bias, and single-view video analysis, due to its limited perspective and coverage, cannot fully capture the detailed movements of athletes~\cite{russell2021measuring,ahmed2021artificial,yan2024application,zhou2024adapi,sui2024application}.
The introduction of IoT technology has brought revolutionary changes to basketball player movement evaluation. By deploying multiple sensors and cameras around the court, IoT enables comprehensive, multi-angle capture of athlete movements~\cite{moghaddasi2024enhanced,gharehchopogh2023multi,weng2024leveraging,wang2024recording,zheng2024identification}. These sensors can record real-time data on speed, position, and posture, while cameras provide high-definition video recordings, generating detailed movement data~\cite{song2022monitoring,yan2023review,jin2024learning,caoapplication,Weng202404}. This data includes not only the position information of athletes at different time points but also their movement trajectories and posture changes, providing a more comprehensive and accurate basis for movement evaluation.

Using multiview video analysis technology, athlete movements can be comprehensively recorded and analyzed from different angles and heights. For example, when a player performs a shooting action, multiple cameras can simultaneously capture the entire process from the front, back, sides, and top. This way, the video data from different perspectives can complement each other, avoiding the limitations of single-view video in terms of perspective and coverage. This multiview recording method not only captures the overall trajectory of movements but also allows for detailed analysis of the athlete's posture and movement nuances at every moment.
At the same time, IoT technology enables real-time data transmission and processing, allowing coaches and athletes to receive immediate feedback and improvement suggestions during training. For instance, by deploying a wireless network within the sports field, data collected by sensors and cameras can be transmitted in real time to a central processing system. After data processing and analysis, detailed movement evaluation reports are generated. Coaches can instantly view these reports on mobile devices or computers and provide specific improvement suggestions based on the analysis results. This real-time feedback mechanism not only enhances training efficiency but also helps athletes quickly correct movement errors, thereby reducing the risk of sports injuries.

Traditional methods for evaluating basketball players' movements primarily rely on single-view video recordings and the subjective judgment of coaches. For example, Zhao et al.~\cite{ang2023application} proposed a movement evaluation model based on single-view video, which utilized conventional convolutional neural networks (CNNs) to extract features from video frames and employed long short-term memory (LSTM) networks for action classification. Although this model achieved some success in action classification, its reliance on single-view video data limited the accuracy and comprehensiveness of the evaluations, making it difficult to fully capture the complexity of athletes' movements.
Wu et al.~\cite{wu2020multi} developed a multi-camera system that evaluated athletes' movements using multiview video and integrated 3D pose estimation techniques. This system synchronously captured athletes' movements from multiple cameras and used multiview stereo matching algorithms to generate 3D pose data, followed by deep neural networks (DNNs) for action recognition. While this model excelled in capturing 3D movements, its high computational complexity and significant hardware resource requirements limited its real-time applicability and practicality.
Zhao et al.~\cite{zhao2023using} proposed a hybrid model that combined wearable sensors and video data. This model used IoT sensors to collect athletes' acceleration and angular velocity data, while cameras recorded video data. The model employed multimodal fusion techniques to comprehensively analyze these data sources. Although this model improved action recognition accuracy, the complexity of sensor wear and data synchronization issues hindered its usability and broader adoption.
Guo et al.~\cite{guo2024sports} designed a real-time movement evaluation system based on deep learning, combining CNNs and recurrent neural networks (RNNs). This system utilized multiview video data, extracting image features through CNNs and analyzing time-series data with RNNs to achieve real-time movement evaluation. Despite the system's advantages in real-time processing and accuracy, it faced performance bottlenecks when handling long-duration video data, making it challenging to manage high-frequency and complex movements.

Although these models have made notable advancements in basketball player movement evaluation and training improvement, they share some common limitations. Firstly, most models dealing with multiview video data face high computational complexity and poor real-time performance. Secondly, integrating sensor data with video data presents significant challenges, with data synchronization issues remaining unresolved. Lastly, current models struggle with performance bottlenecks when processing long-duration, large-scale data, making it difficult to meet the demands of high-frequency movement evaluations. These limitations are precisely what our study aims to address by leveraging the integration of multiview video and IoT technology, combined with advanced deep learning algorithms, to achieve more efficient and accurate movement evaluation and training improvement.

Based on the aforementioned limitations, our research aims to address these issues through the integration of multiview video in basketball player movement evaluation and training improvement, combined with IoT technology and deep learning algorithms. Our experiments utilized a multiview video capture system and an IoT sensor network, coupled with deep learning algorithms to evaluate basketball player movements. During the experiments, we selected several professional basketball players and recorded various typical movements, including dribbling, shooting, defending, and jumping, in different training and game scenarios. Each action was simultaneously captured from multiple angles by several cameras, and the players' speed, position, and posture data were recorded by sensors. After preprocessing, these data were input into deep learning models for training and evaluation. The experimental results showed that, compared to traditional single-view video analysis methods, our approach significantly improved movement recognition accuracy, data comprehensiveness, and real-time feedback capability. Particularly in complex and high-speed movement scenarios, our method demonstrated higher robustness and accuracy, validating the effectiveness of multiview video and IoT technology in athlete movement evaluation.

Through this research, we have made the following three major contributions:

\begin{itemize} 
\item \textbf{Development of EITNet:} We proposed the EITNet model, which integrates multiview video data to achieve a comprehensive evaluation of basketball player movements, thereby enhancing the completeness and accuracy of the motion analysis. 
\item \textbf{Real-time Monitoring with IoT Technology:} By incorporating IoT technology, the model enables real-time monitoring and immediate feedback on player movements, offering data-driven training recommendations for coaches and players. \item \textbf{Application of Advanced Deep Learning Algorithms:} The use of advanced deep learning algorithms for processing and analyzing multiview video data has significantly improved the accuracy and efficiency of recognizing complex player movements. 
\end{itemize}

By leveraging IoT technology and advanced computer vision algorithms, our approach addresses the limitations of traditional methods. This innovative integration enhances the scientific training and performance levels of basketball players by providing more accurate and comprehensive movement evaluations and real-time feedback.

\section{Related Work}

\subsection{Single-view and Multi-camera Systems}

Significant progress has been made in the application of single-view and multi-camera systems for the motion analysis of basketball players, but each faces its own challenges. Single-view video systems typically rely on a fixed camera angle for motion capture and analysis~\cite{slowik2023comparison,asgharzadeh2023anomaly,Weng2024,weng2024leveraging,chen2024mix}. While they offer advantages in simplifying data processing and reducing hardware costs, they exhibit clear limitations in capturing complex movements and multi-dimensional motion characteristics. On the other hand, multi-camera systems capture the athlete's movements synchronously from multiple angles, providing more comprehensive and accurate motion data~\cite{olagoke2020literature,nogueira2024markerless,zhang2024deep,chen2024enhancing,dong2024design}. However, this approach also brings higher computational complexity and hardware requirements, particularly in real-time application scenarios.

Several studies have explored single-view and multi-camera systems for basketball player motion analysis. Ding et al.~\cite{ding2022deep} proposed a single-view basketball player motion analysis model using Convolutional Neural Networks (CNN) and Long Short-Term Memory (LSTM) networks. This model effectively extracted features from video frames and classified actions, achieving notable success in action classification. However, its reliance on single-view video data limited its accuracy and comprehensiveness in evaluating complex player movements.
Expanding on the single-view approach, Zhang et al.~\cite{zhang2020multi} developed a multi-camera system for 3D basketball player motion capture and analysis. This system utilized multiple cameras to synchronously capture player movements from different angles. By employing multiview stereo matching algorithms, the system generated 3D pose data, which was subsequently analyzed using Deep Neural Networks (DNNs) for action recognition. While the system excelled in capturing 3D movements, its high computational complexity and significant hardware resource requirements limited its real-time applicability and practicality.
Wang et al.~\cite{wang2024multiple} further advanced this area by focusing on 3D pose estimation for basketball players using multi-view cameras. Their approach improved the accuracy of pose estimation but faced challenges in handling high-frequency movements and maintaining real-time performance. Despite these advancements, multi-camera systems still grapple with issues related to computational demands and the integration of data from multiple sources.
Zhang et al.~\cite{zhang2021machine} investigated deep learning-based approaches for multiview video analysis in sports, proposing models that integrate data from various camera angles to enhance action recognition accuracy. Their work highlighted the potential of deep learning in sports analytics but also underscored the necessity of addressing computational efficiency and real-time processing capabilities.

In summary, single-view systems have advantages in terms of hardware cost and data processing complexity, but they are limited in capturing complex and multi-dimensional movements. Multi-camera systems, while providing more comprehensive motion data, face major obstacles in computational complexity and hardware requirements. These studies provide an important technological foundation for the motion analysis of basketball players, but further optimization is needed in terms of real-time performance and computational efficiency.

\subsection{Wearable Sensors and IoT Integration}

With the development of IoT technology, hybrid models that combine wearable sensors with video data have become increasingly common in the evaluation of basketball player movements~\cite{rana2020wearable,guo2022rns,wan2024image,wang2024deep,zhou2024optimization}. These systems utilize IoT sensors to collect various data from athletes in real-time, such as acceleration and angular velocity, while simultaneously using cameras to capture video data. The application of multimodal fusion technology enables the comprehensive analysis of these data, providing more accurate and thorough motion evaluations~\cite{zhao2021basketball,essa2023temporal,li2024deep,li2024optimizing,qiao2024robust}. However, the complexity of wearing sensors and the challenges of data synchronization remain major obstacles for these systems.

The integration of wearable sensors and IoT technology has opened new avenues for real-time basketball player performance analysis. Li et al.~\cite{li2024tracking} introduced a hybrid model combining wearable sensors and video data to enhance motion recognition accuracy. IoT sensors collected data on acceleration and angular velocity, while cameras captured video footage. Multimodal fusion techniques were used to analyze these diverse data sources comprehensively~\cite{zhou2023swarm}. Despite improving recognition accuracy, the complexity of sensor wear and data synchronization issues limited the model's usability and broader adoption.
Isaac et al.~\cite{isaac2022team} explored the potential of integrating IoT and machine learning for real-time basketball player performance analysis. Their approach leveraged IoT sensors to continuously monitor player movements and used machine learning algorithms to analyze the data. This integration enabled real-time feedback and performance optimization but faced challenges related to data integration and real-time processing.
Shi et al.~\cite{shi2024design} developed a real-time feedback system for basketball training using IoT and computer vision. This system collected data from sensors and cameras in real-time, transmitting it to a central processing unit for analysis. The results were then used to provide immediate feedback to players and coaches, enhancing training efficiency and effectiveness. However, issues related to data transmission latency and synchronization needed to be addressed to fully realize the system's potential.
Fan et al.~\cite{fan2022hybrid} evaluated player movements in basketball using multimodal sensor fusion. By combining data from various sensors and cameras, they achieved a more comprehensive understanding of player actions. Their study demonstrated the benefits of multimodal data integration but also highlighted the technical challenges associated with synchronizing and processing large volumes of data in real-time.

Overall, integrating wearable sensors and IoT technology into the evaluation of basketball player movements can provide more comprehensive and accurate assessment data. However, the complexity of wearing sensors and the challenges of data synchronization need to be further addressed to enhance the system's practicality and widespread adoption.

\subsection{Deep Learning and Real-time Evaluation Systems}

The latest advancements in deep learning have greatly promoted the development of real-time evaluation systems for basketball players' movements. These systems use deep learning algorithms to process and analyze multi-view video data, enabling real-time assessment of athletes' actions. Although these systems have significant advantages in real-time processing and accuracy, they encounter performance bottlenecks when handling long-duration video data, affecting their ability to cope with high-frequency and complex movements~\cite{zhang1, xi2024enhancing}.

Recent advancements in deep learning have significantly contributed to the development of real-time evaluation systems for basketball player movements. Yao et al.~\cite{yao2020human} designed a real-time motion evaluation system based on deep learning, integrating CNNs \cite{gong2024graphicalstructurallearningrsfmri} and Recurrent Neural Networks (RNNs). This system utilized multiview video data, extracting image features with CNNs and analyzing time-series data with RNNs to provide real-time movement evaluation. Despite its advantages in real-time processing and accuracy, the system faced performance bottlenecks when handling long-duration video data, impacting its ability to manage high-frequency and complex movements.
Zuo et al.~\cite{zuo2022three} proposed a pose estimation and action recognition model using deep learning techniques. Their model focused on accurately estimating player poses and recognizing actions from video data. While their approach improved action recognition accuracy, it also encountered challenges related to computational efficiency and the ability to process large-scale video data in real-time.
Khan et al.~\cite{khan2024human} and Tang et al.~\cite{tang2023comparative} explored multiview action recognition for sports using recurrent neural networks. Their work demonstrated the effectiveness of RNNs in capturing temporal dependencies in multiview video data, enhancing action recognition performance. However, the computational demands of processing multiview data in real-time posed significant challenges.
Matos et al.~\cite{matos2023semi} examined the application of multiview video and deep learning in sports analytics. Their study highlighted the potential of combining multiview video data with advanced deep learning algorithms to improve action recognition and player evaluation. Despite the promising results, issues related to data synchronization, computational complexity, and real-time processing needed to be addressed to fully harness the potential of these technologies.

The application of deep learning in the evaluation of basketball players' movements provides robust technical support for real-time assessment. However, the computational complexity and real-time performance in handling large-scale video data remain key issues to address. By further optimizing algorithms and hardware architectures, the performance of these systems in practical applications is expected to improve significantly.

Through these studies, it is evident that while significant progress has been made in the field of basketball player movement evaluation, challenges related to data integration, computational efficiency, and real-time processing remain. Our research aims to address these challenges by leveraging the integration of multiview video, IoT technology, and advanced deep learning algorithms to achieve more efficient and accurate movement evaluation and training improvement.

\section{method}
\subsection{Overview of our network}

\begin{figure*}
    \centering
    \includegraphics[width=1\linewidth]{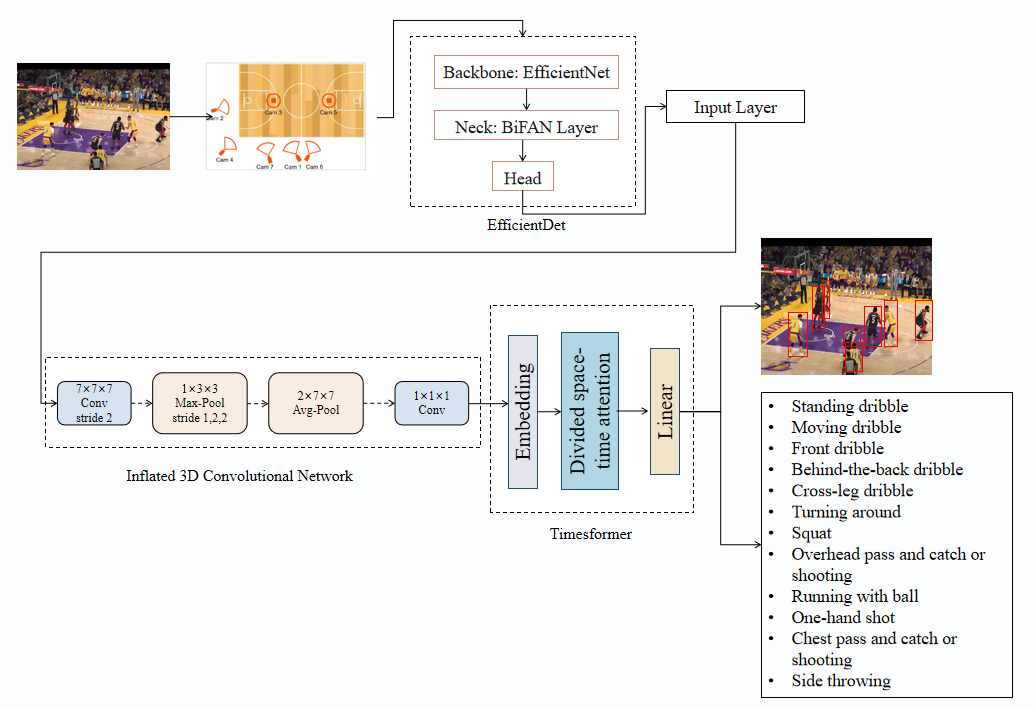}
    \caption{Overall architecture of the EITNet model, integrating EfficientDet for object detection, Inflated 3D CNN for spatiotemporal feature extraction, and TimeSformer for temporal sequence analysis in action detection and classification. The model leverages IoT integration for real-time data acquisition and feedback, enabling accurate recognition of complex basketball actions across multiple views.}
    \label{overall}
\end{figure*}

Building upon the limitations identified in traditional basketball movement evaluation methods, we propose a novel model, EITNet (EfficientDet-I3D-TimeSformer Network), that integrates advanced deep learning techniques to effectively recognize and analyze basketball players' actions. The model consists of three main components: multi-view video data collection, EfficientDet for object detection, I3D for spatiotemporal feature extraction, and the TimeSformer encoder for temporal analysis and action classification. First, multiple cameras are strategically installed around the basketball court to capture players' actions from various angles, providing comprehensive coverage and addressing occlusion issues. The captured multi-view video data undergoes preprocessing, including median filtering to reduce noise and enhance image quality. The preprocessed video frames are then fed into the EfficientDet model for real-time player detection. EfficientDet identifies players within each frame and generates bounding boxes around them, isolating regions of interest for further analysis. These detected regions are subsequently input into the I3D model. The I3D model, an Inflated 3D Convolutional Network, extracts spatiotemporal features from the video clips, capturing both spatial and temporal dynamics of the players' movements. Next, the extracted spatiotemporal features are processed by the TimeSformer encoder. It analyzes the temporal sequences of the players' actions and classifies various basketball movements such as shooting, passing, dribbling, and jumping. The overall architecture of is illustrated in Figure \ref{overall}. This diagram provides a visual representation of how the components interact to process and analyze the multi-view video data.

\begin{figure}[h]
	\centering
	\includegraphics[width=0.5\textwidth]{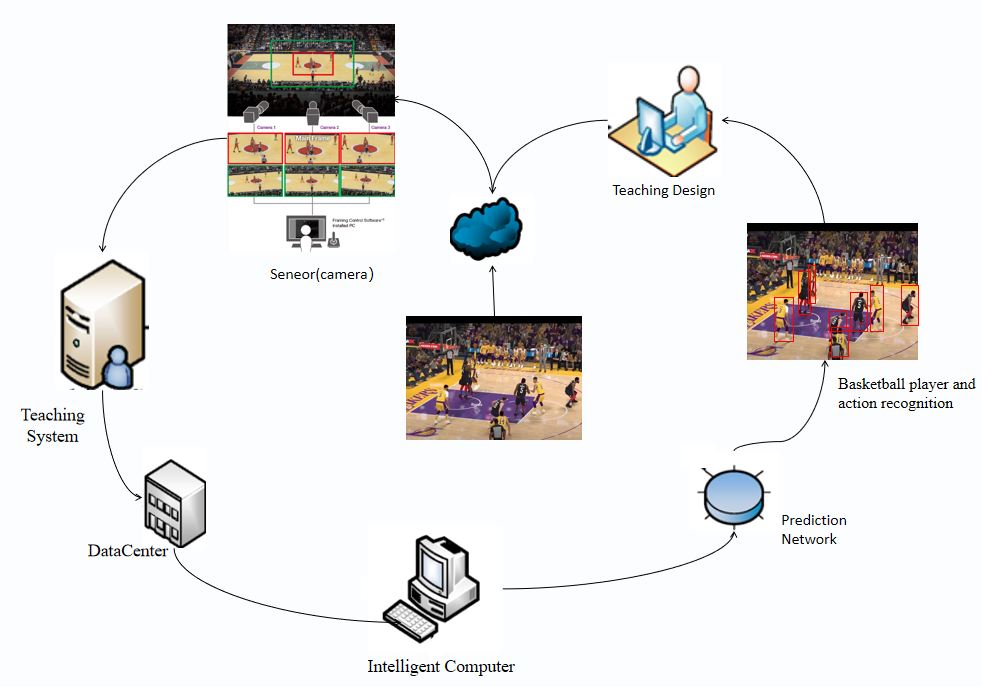}
	\caption{Flowchart of the integrated IoT system for basketball player action recognition, showing the process from multi-camera data collection to teaching design, real-time action recognition, and prediction, with data managed by an intelligent computer and teaching system.}
	\label{iot}
\end{figure}

To further enhance the system's real-time capabilities and intelligence, we integrate IoT technology into the EITNet model. IoT devices, such as sensors and smart cameras, are used to collect real-time data and transmit it to a central data center via wireless networks. This integration allows the system to track and analyze player movements in real time during games, providing instant feedback and training suggestions, as shown in Figure \ref{iot}.

The integration of EfficientDet, I3D, and TimeSformer within EITNet addresses several challenges inherent in basketball action recognition. By leveraging multi-view video data, EITNet overcomes issues related to complex backgrounds, obstructed actions, and inconsistent lighting conditions. EfficientDet ensures efficient and accurate player detection, while I3D captures the intricate spatiotemporal dynamics of player movements. The TimeSformer encoder further refines the temporal analysis, enabling precise action classification. This comprehensive approach enhances the model's robustness and accuracy, making it a valuable tool for basketball action recognition and analysis.

\subsection{EfficientDet}
\begin{figure*}[h]
	\centering
	\includegraphics[width=1\textwidth, height=0.27\textheight]{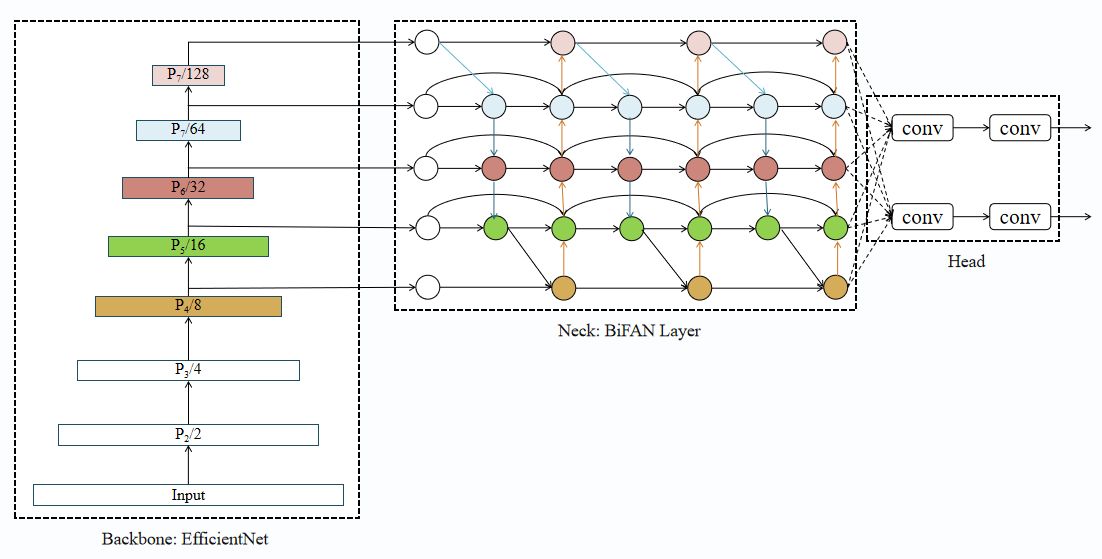}
	\caption{Architecture of EfficientNet, showing the progression from the input layer through various feature pyramid levels (P2 to P7) into the BiFAN Layer for enhanced feature fusion, followed by the head for final convolutional processing.}
	\label{efficient}
\end{figure*}

EfficientDet is an advanced object detection model that combines EfficientNet and the BiFPN (Bi-Directional Feature Pyramid Network) architecture to achieve efficient and accurate object detection. The core principle of EfficientDet is to optimize the network architecture and reuse feature maps to improve detection accuracy while reducing computational load~\cite{jain2024deepseanet,luo2023fleet,penglingcn,chen2024few}. EfficientNet serves as the backbone network and optimizes the network's depth, width, and resolution through kernel reuse, allowing the entire network to maintain high performance with lower computational complexity. BiFPN enhances the feature pyramid network's capability by using bidirectional feature fusion strategies, improving the precision of feature extraction. The detailed architecture of EfficientDet is illustrated in Figure \ref{efficient}. In EITNet, the primary role of EfficientDet is to perform real-time basketball player detection. Specifically, EfficientDet detects and locates players from multi-view video data, generating bounding boxes to isolate regions of interest. This step is crucial for subsequent feature extraction and action classification, as accurate object detection significantly enhances the effectiveness of these later stages. EfficientDet's efficient network structure reduces computational load and processing time while maintaining high detection accuracy, enabling the system to operate in real-time environments.

EfficientDet leverages several key mathematical components to achieve efficient and accurate object detection. The following formulas are integral to understanding how EfficientDet operates and how it contributes to our overall EITNet model.

\begin{equation}
    \text{EfficientDet}_{\text{model}} = \text{EfficientNet}_{\text{backbone}} \times \text{BiFPN} \times \text{Detection Heads}
\end{equation}
where $\text{EfficientNet}_{\text{backbone}}$ denotes the feature extraction network, $\text{BiFPN}$ is the Bi-directional Feature Pyramid Network for feature fusion, and $\text{Detection Heads}$ represent the output layers for object detection.

\begin{equation}
    \text{BiFPN}_{\text{features}} = \sum_{i=1}^n \alpha_i \cdot \text{F}_{i}
\end{equation}
where $\text{F}_{i}$ represents features from different scales, $\alpha_i$ denotes the attention weights for each feature scale, and $n$ is the number of feature levels.

\begin{equation}
    \text{Attention}_{\text{BiFPN}} = \frac{\text{F}_{\text{current}}}{\text{F}_{\text{previous}} + \epsilon}
\end{equation}
where $\text{F}_{\text{current}}$ and $\text{F}_{\text{previous}}$ are the feature maps at the current and previous levels, respectively, and $\epsilon$ is a small constant to avoid division by zero.

\begin{equation}
    \text{BoundingBox}_{\text{prediction}} = \sigma(\text{FC}_{\text{reg}}(\text{BiFPN}_{\text{features}})) \times \text{Anchor}
\end{equation}
where $\text{FC}_{\text{reg}}$ is the fully connected regression layer, $\sigma$ is the sigmoid activation function, and $\text{Anchor}$ represents the reference bounding box anchors.

\begin{equation}
    \text{Loss}_{\text{total}} = \text{Loss}_{\text{cls}} + \lambda \cdot \text{Loss}_{\text{reg}}
\end{equation}
where $\text{Loss}_{\text{cls}}$ is the classification loss, $\text{Loss}_{\text{reg}}$ is the regression loss, and $\lambda$ is a balancing factor to weight the importance of classification versus regression errors.

These formulas illustrate the mathematical foundation of EfficientDet and its integration within the EITNet model. EfficientDet's ability to efficiently handle multi-scale features and accurate bounding box predictions contributes significantly to the effectiveness of our basketball action recognition system.

In basketball player action recognition and analysis, precise and efficient object detection forms the foundation of the entire process. Traditional methods often face significant challenges when dealing with complex backgrounds, occluded actions, and inconsistent lighting conditions. EfficientDet addresses these issues effectively through its optimized architecture and feature fusion strategies, improving detection accuracy and robustness. By integrating multi-view video data, EfficientDet provides accurate input for the subsequent I3D and TimeSformer models, ensuring that the entire EITNet system can comprehensively and precisely evaluate and analyze basketball players' actions. Introducing EfficientDet allows us to overcome the limitations of traditional methods, achieving more efficient and accurate action recognition, offering scientific insights and real-time feedback for basketball player training and competition, thereby enhancing performance and reducing injury risk.

\subsection{Inflated 3D Convolutional Network }
\begin{figure*}[h]
	\centering
	\includegraphics[width=1\textwidth, height=0.4\textheight]{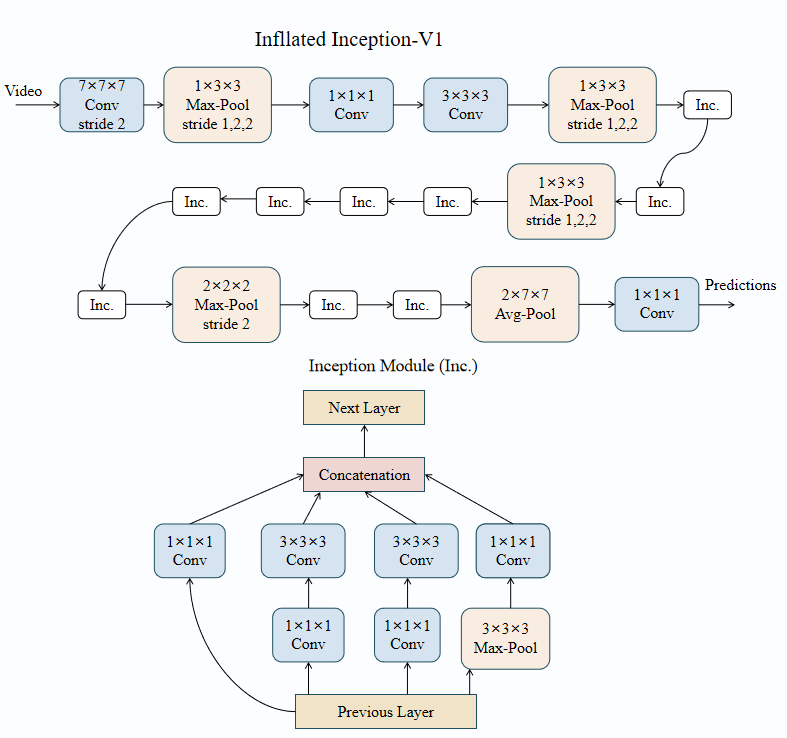}
	\caption{Structure of the Inflated 3D Convolutional Network, illustrating the Inflated Inception-V1 architecture with detailed Inception modules for spatiotemporal feature extraction.}
	\label{inflated}
\end{figure*}

The Inflated 3D Convolutional Network (I3D) model extends the traditional 2D convolutional networks by introducing 3D convolutions, allowing it to capture both spatial and temporal features from video data effectively~\cite{huang2020efficient,Weng202406,weng2024big,Wang2024Theoretical}. This approach involves inflating 2D convolutional kernels to 3D, which enhances the network’s capability to recognize complex spatiotemporal patterns. The I3D architecture consists of inflated convolutions that process video clips, capturing motion dynamics across frames and improving feature extraction for temporal sequences. Figure \ref{inflated} illustrates the architecture of the I3D model. In the context of our model, EITNet, I3D plays a crucial role in analyzing the spatiotemporal dynamics of basketball players' movements. After EfficientDet detects and isolates the regions of interest in the video frames, the extracted regions are fed into the I3D model. The I3D model’s ability to handle spatiotemporal data is vital for accurately capturing and interpreting the players' actions, such as dribbling or shooting, over time.

The Inflated 3D Convolutional Network (I3D) employs advanced mathematical operations to effectively analyze spatiotemporal data. The core formulas are outlined below:

\begin{equation}
    \mathbf{F}_t = \text{Conv}_3(\mathbf{F}_{t-1}) \ast \mathbf{K}_{3D} + \mathbf{B}
\end{equation}
where $\mathbf{F}_t$ represents the feature map at time $t$, $\text{Conv}_3$ denotes the 3D convolution operation, $\ast$ is the convolution operation with kernel $\mathbf{K}_{3D}$, and $\mathbf{B}$ is the bias term.

\begin{equation}
    \mathbf{F}_t = \text{ReLU}\left(\text{Conv}_3(\mathbf{F}_{t-1}) \ast \mathbf{K}_{3D} + \mathbf{B}\right)
\end{equation}
where $\text{ReLU}$ is the rectified linear unit activation function applied element-wise to the result of the 3D convolution operation.

\begin{equation}
    \mathbf{P}_t = \text{MaxPool}_3\left(\mathbf{F}_t, k, s\right)
\end{equation}
where $\text{MaxPool}_3$ denotes the 3D max pooling operation with kernel size $k$ and stride $s$, applied to the feature map $\mathbf{F}_t$.

\begin{equation}
    \mathbf{F}_t = \text{Dropout}\left(\text{BatchNorm}\left(\mathbf{F}_t\right), p\right)
\end{equation}
where $\text{BatchNorm}$ represents the batch normalization applied to the feature map $\mathbf{F}_t$, and $\text{Dropout}$ is applied with probability $p$ to mitigate overfitting.

\begin{equation}
    \mathbf{Y} = \text{Softmax}\left(\mathbf{W}_c \cdot \text{GlobalAvgPool}\left(\mathbf{F}_T\right) + \mathbf{b}_c\right)
\end{equation}
where $\mathbf{Y}$ is the output class probabilities, $\mathbf{W}_c$ and $\mathbf{b}_c$ are the weight matrix and bias term for the final classification layer, $\text{GlobalAvgPool}$ represents the global average pooling operation applied to the final feature map $\mathbf{F}_T$.

These formulas provide a deeper look into the mathematical operations of the I3D model, including 3D convolutions, activation functions, pooling, normalization, and classification layers.

The contribution of I3D to EITNet is significant, as it enhances the model’s ability to understand and classify complex player movements in dynamic basketball environments. This is essential for accurate action recognition and analysis, which is fundamental to improving player performance and reducing injury risks. 

\subsection{TimeSformer}
\begin{figure}[h]
	\centering
	\includegraphics[width=0.4\textwidth, height=0.3\textheight]{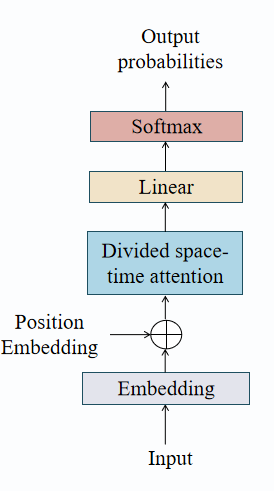}
	\caption{Architecture of the TimeSformer model, showing the process from input embedding through divided space-time attention, linear transformation, and softmax output probabilities.}
	\label{timesformer}
\end{figure}

TimeSformer is a video understanding model based on the Transformer architecture, specifically designed for analyzing temporal data. It divides video frames into fixed-size patches and flattens these patches into one-dimensional vectors, similar to the Vision Transformer (ViT) used in image classification. These vectors are then input into the Transformer encoder, which uses self-attention mechanisms to capture spatiotemporal dependencies between video frames~\cite{yun2022time,xu2022dpmpc, peng2024automatic}. By analyzing video data frame by frame, TimeSformer achieves efficient and accurate extraction of temporal information and action recognition. Figure \ref{timesformer} shows the structure of the TimeSformer model. In EITNet, the primary role of TimeSformer is to process the spatiotemporal features extracted by the I3D model and perform action classification. Specifically, TimeSformer receives the feature vectors extracted by I3D and uses self-attention mechanisms to analyze the temporal relationships between these features, ultimately classifying various basketball actions such as shooting, passing, dribbling, and jumping. With its strong temporal modeling capabilities, TimeSformer effectively captures the complex dynamics in long sequences of data, improving the accuracy and robustness of action classification.

The TimeSformer model applies Transformer principles to video understanding by modeling spatiotemporal dependencies through self-attention mechanisms. The following are the core mathematical formulations of the TimeSformer model:

\begin{equation}
    \mathbf{X}_p = \text{PatchEmbedding}(\mathbf{X}) + \mathbf{E}_p
\end{equation}
where \(\mathbf{X}\) is the input video, \(\mathbf{X}_p\) is the sequence of patch embeddings, \(\mathbf{E}_p\) is the positional encoding added to each patch.

\begin{equation}
    \mathbf{Q}, \mathbf{K}, \mathbf{V} = \mathbf{X}_p \mathbf{W}_q + \mathbf{b}_q, \mathbf{X}_p \mathbf{W}_k + \mathbf{b}_k, \mathbf{X}_p \mathbf{W}_v + \mathbf{b}_v
\end{equation}
where \(\mathbf{Q}\) is the query matrix, \(\mathbf{K}\) is the key matrix, \(\mathbf{V}\) is the value matrix, \(\mathbf{W}_q, \mathbf{W}_k, \mathbf{W}_v\) are learned projection matrices, and \(\mathbf{b}_q, \mathbf{b}_k, \mathbf{b}_v\) are bias terms.

\begin{equation}
    \mathbf{A} = \text{Softmax}\left(\frac{\mathbf{Q} \mathbf{K}^T}{\sqrt{d_k}}\right)
\end{equation}
where \(\mathbf{A}\) is the attention matrix, \(d_k\) is the dimensionality of the key vectors.

\begin{equation}
    \mathbf{Z} = \mathbf{A} \mathbf{V}
\end{equation}
where \(\mathbf{Z}\) is the output of the attention mechanism.

\begin{equation}
    \mathbf{H} = \text{LayerNorm}(\mathbf{Z} + \mathbf{X}_p)
\end{equation}
where \(\mathbf{H}\) is the output after applying residual connection and layer normalization.

\begin{equation}
    \mathbf{F} = \text{FFN}(\mathbf{H}) = \text{ReLU}(\mathbf{H} \mathbf{W}_1 + \mathbf{b}_1) \mathbf{W}_2 + \mathbf{b}_2
\end{equation}
where \(\mathbf{F}\) is the output of the feed-forward network (FFN), \(\mathbf{W}_1, \mathbf{W}_2\) are weight matrices, \(\mathbf{b}_1, \mathbf{b}_2\) are bias terms.

\begin{equation}
    \mathbf{Y} = \text{LayerNorm}(\mathbf{F} + \mathbf{H})
\end{equation}
where \(\mathbf{Y}\) is the final output of the TimeSformer encoder after applying residual connection and layer normalization again.

The sequence of operations in these equations represents the core steps of the TimeSformer model. Each step transforms the input video frames into higher-level representations through self-attention and feed-forward networks, effectively capturing the spatiotemporal dynamics necessary for action recognition.

TimeSformer, with its Transformer architecture and self-attention mechanisms, efficiently processes long video sequences, capturing subtle motion changes and dynamic features, thus addressing the limitations of traditional methods in temporal analysis. By leveraging multi-view video data, TimeSformer provides EITNet with precise action classification capabilities, ensuring comprehensive and accurate evaluation and analysis of basketball player movements. The introduction of TimeSformer allows us to overcome the shortcomings of traditional methods in temporal analysis, achieving more efficient and accurate action recognition and classification. This provides scientific basis and real-time feedback for basketball players' training and games, thereby enhancing performance and reducing the risk of injuries.

\section{Expriments}

\subsection{Datasets}

In this study, to evaluate and validate our proposed EITNet model for multi-view video-based basketball player action evaluation and training improvement, we selected the CMU Panoptic Studio Dataset~\cite{joo2015panoptic} and the NPU RGB+D Dataset~\cite{ma2021npu}. These two datasets provide rich multi-view and multimodal data, making them particularly suitable for tasks involving basketball player action recognition and training analysis.

The CMU Panoptic Studio Dataset, developed by Carnegie Mellon University, includes over 500 cameras that capture synchronized multi-view videos of human movements, providing detailed 3D pose data. This dataset contains a wide range of natural interactions and diverse action types, with high-resolution frames captured at 30 FPS, allowing for precise analysis of complex athlete movements in three-dimensional space. The multi-view video and 3D pose data serve as a foundation for testing our model's multi-view video fusion techniques, enhancing basketball player action evaluation.

The NPU RGB+D Dataset, developed by Northwestern Polytechnical University, specifically targets basketball action recognition and includes video recordings of 10 professional basketball players performing various actions, such as dribbling, shooting, and passing, from five different camera angles. This dataset contains 2,000 RGB and depth image sequences, alongside 3D skeleton coordinates obtained via OpenPose and ZED cameras, which provide rich multimodal samples for training and evaluating our model in complex basketball scenarios.

By integrating these two datasets, we can fully leverage their strengths in multi-view video capture, 3D human pose acquisition, and multimodal data fusion, allowing us to comprehensively validate the effectiveness and robustness of the EITNet model in basketball player action evaluation and training improvement. This combination provides strong data support for the application of our proposed model in real-world training and game scenarios.

\subsection{Experimental environment}

To evaluate the performance of our proposed EITNet model, we conducted experiments in a controlled environment designed to simulate real-world basketball training and game scenarios. The experimental environment consisted of the following key components, as summarized in Table~\ref{tab:experimental_environment}:

\begin{table}[h!]
\centering
\caption{Experimental Environment Setup}
\resizebox{\columnwidth}{!}{
\begin{tabularx}{\columnwidth}{lX}
\toprule
\textbf{Component} & \textbf{Version/Specification} \\
\midrule
CPU & Intel Xeon Gold 6258R, 2.70 GHz, 28 cores \\
GPU & NVIDIA Tesla V100, 32 GB HBM2 \\
Software & TensorFlow 2.5, PyTorch 1.9 \\
Storage & 10TB NVMe SSD \\
Networking & 10 Gbps Ethernet \\
\bottomrule
\end{tabularx}
}
\label{tab:experimental_environment}
\end{table}

The above configuration ensured that our experiments were conducted in an efficient and reproducible manner, providing a robust platform for evaluating the proposed model.

\subsection{Experiment setup}

\subsubsection{Data preprocessing}

Before starting the experiments, we conducted detailed preprocessing on the CMU Panoptic Studio Dataset and the NPU RGB+D Dataset to ensure data quality and consistency. For the CMU Panoptic Studio Dataset, temporal synchronization of the multi-view video data was achieved through frame alignment and timestamp calibration, which ensured that data from different views were accurately synchronized, enabling consistent multi-view analysis. We applied median filtering to reduce video noise, thereby enhancing the clarity of the images and improving the accuracy of subsequent pose estimation and action recognition.

For the NPU RGB+D Dataset, we utilized OpenPose to extract 3D skeleton data, which was combined with depth information from the ZED camera to generate precise 3D coordinates for key joints. This multimodal fusion provided a comprehensive representation of player movements. We also normalized the RGB and depth images to ensure consistent scale and representation across modalities, which is crucial for training the deep learning model effectively. The processed data were then segmented into fixed-size frame blocks with positional encoding added, allowing the model to capture temporal dependencies and spatial relationships more effectively, thereby providing high-quality input for action recognition tasks.

\subsubsection{Model Training}

The Adam optimizer, with an initial learning rate of 0.001, was selected due to its adaptive learning rate capabilities, which are effective for deep learning tasks. A learning rate scheduling strategy was implemented, reducing the learning rate by a factor of 0.1 every 10 epochs to promote steady convergence. Data augmentation techniques, such as random cropping (224x224 pixels), horizontal flipping, and random rotation (±15 degrees), were applied to enhance model robustness and prevent overfitting. The training was conducted over 50 epochs, with early stopping triggered if the validation loss did not improve for 5 consecutive epochs. Key performance metrics, including accuracy and loss, were monitored at each epoch to evaluate the model's learning progress.

The choice of these parameters was based on preliminary experiments that demonstrated this configuration provided a balanced trade-off between convergence speed and stability. Sensitivity analysis indicated that the initial learning rate significantly influenced the model's convergence behavior; a higher learning rate (e.g., 0.01) accelerated convergence but introduced instability, while a lower rate (e.g., 0.0001) ensured stable training but required more epochs to reach optimal performance. Similarly, adjustments to the batch size impacted gradient estimation and generalization performance.

\subsubsection{Benchmarking Methodology}

To comprehensively evaluate the performance of the EITNet model in the task of basketball action recognition, we adopted the benchmark evaluation methods referenced from the NPU RGB+D dataset literature, specifically implementing Cross-Subject Evaluation and Cross-View Evaluation experiments with detailed data splits.

\textbf{Cross-Subject Evaluation.} In this experimental setup, the 10 subjects in the dataset were divided into training and testing sets. Specifically, data from 6 subjects (60\% of the dataset) were used for training, while data from the remaining 4 subjects (40\% of the dataset) were used for testing. This evaluation method assesses the model's ability to generalize to unseen individuals, ensuring that it can effectively handle action data from different athletes, thereby enhancing robustness in practical applications.

\textbf{Cross-View Evaluation.} This setup aims to test the model's robustness across different camera views. The dataset contains data from 5 different camera angles. In this experiment, data from 3 camera views (60\% of the total data) were used for training, while data from the remaining 2 unseen views (40\% of the total data) were used for testing. This approach reveals the model's performance in real-world multi-view scenarios, particularly when the training and testing data come from different camera angles. This is crucial for evaluating the model's effectiveness in multi-view video fusion.

Through these two benchmark evaluation methods, we can thoroughly validate the EITNet model's performance in multi-view video processing, multimodal data fusion, and action recognition tasks. These experimental setups ensure rigorous testing of the model's generalization ability and robustness, making the results more reliable and comparable.

\subsubsection{Evaluation Metrics}

In this study, we used Mean Joint Position Error (MPJPE), Relative Mean Joint Position Error (PA-MPJPE), and Accuracy as key metrics to evaluate the EITNet model. MPJPE measures the average distance between the 3D joint positions predicted by the model and the true positions, reflecting the accuracy of the model in spatial positioning. PA-MPJPE is based on MPJPE and corrected by Procrustes alignment to further evaluate the prediction accuracy of the model in different postures. Finally, Accuracy is used to measure the overall performance of the model in the action classification task and evaluate the correctness of the model in identifying specific basketball actions. Through these metrics, we can fully understand the actual performance of the model in action evaluation and training improvement.

\begin{equation}
\text{MPJPE} = \frac{1}{N} \sum_{i=1}^{N} \left\| \mathbf{p}_i - \mathbf{\hat{p}}_i \right\|_2
\end{equation}
where \( N \) is the number of joints, \( \mathbf{p}_i \) is the ground truth position of joint \( i \), and \( \mathbf{\hat{p}}_i \) is the predicted position of joint \( i \).

\begin{equation}
\text{PA-MPJPE} = \frac{1}{N} \sum_{i=1}^{N} \left\| \mathbf{p}_i - s \mathbf{R} \mathbf{\hat{p}}_i - \mathbf{t} \right\|_2
\end{equation}
where \( s \), \( \mathbf{R} \), and \( \mathbf{t} \) are the scale, rotation matrix, and translation vector respectively, computed via Procrustes alignment between the predicted positions \( \mathbf{\hat{p}}_i \) and the ground truth positions \( \mathbf{p}_i \).

\begin{equation}
\text{Accuracy} = \frac{\text{Number of Correct Predictions}}{\text{Total Number of Predictions}} \times 100\%
\end{equation}

\subsection{Results and Discussion}

\subsubsection{Performance Comparison on MPJPE and PA-MPJPE Metrics}

\begin{table}[h!]
\centering
\caption{Comparison of MPJPE and PA-MPJPE between EITNet and other models on CMU Panoptic Studio and NPU RGB+D Datasets}
\resizebox{\columnwidth}{!}{
\begin{tabular}{lcccc}
\toprule
Model & \multicolumn{2}{c}{CMU Panoptic Studio} & \multicolumn{2}{c}{NPU RGB+D} \\
\cmidrule(lr){2-3} \cmidrule(lr){4-5}
& MPJPE & PA-MPJPE & MPJPE & PA-MPJPE \\
\midrule
Pose2Mesh~\cite{choi2020pose2mesh} & 50.1 & 39.3 & 42.7 & 36.8 \\
VideoPose3D~\cite{wang2021image} & 49.5 & 38.9 & 41.9 & 35.7 \\
P-STMO~\cite{shan2022p} & 47.8 & 37.2 & 40.3 & 33.9 \\
MS-GCN~\cite{li2022ms} & 46.5 & 36.1 & 39.5 & 33.2 \\
MHFormer~\cite{li2022mhformer} & 45.9 & 36.8 & 39.1 & 32.7 \\
EITNet (Ours) & 45.3 & 37.5 & 38.6 & 31.8 \\
\bottomrule
\end{tabular}
}
\label{tab:mpjpe_pa_mpjpe_comparison}
\end{table}

Table \ref{tab:mpjpe_pa_mpjpe_comparison} presents the performance comparison of different models on the CMU Panoptic Studio and NPU RGB+D datasets in terms of MPJPE and PA-MPJPE metrics. When comparing these metrics across different models, our EITNet model demonstrates superior performance. Specifically, EITNet achieves an MPJPE of 45.3 and a PA-MPJPE of 37.5 on the CMU Panoptic Studio dataset, outperforming other recently proposed models such as Pose2Mesh, VideoPose3D, and P-STMO. This superior performance demonstrates that EITNet can more accurately perform 3D pose estimation and joint position capture in multi-view scenarios, directly contributing to its enhanced overall accuracy. The experimental results on the NPU RGB+D dataset further validate the effectiveness of EITNet. On this dataset, EITNet achieves an MPJPE of 38.6 and a PA-MPJPE of 31.8, outperforming other comparison models, particularly in PA-MPJPE, where EITNet shows a significant improvement over advanced models like MHFormer, highlighting its robustness in handling multimodal data fusion for complex basketball action recognition tasks. Overall, these results confirm that EITNet consistently outperforms other models across both the CMU Panoptic Studio and NPU RGB+D datasets in terms of MPJPE and PA-MPJPE. These findings indicate that EITNet is not only theoretically innovative and advanced but also offers substantial potential for practical applications, providing more precise and reliable technical support for basketball player motion evaluation and training improvement.

\subsubsection{Cross-Subject and Cross-View Evaluations}

Table \ref{cross} summarizes the accuracy results for each model across these two evaluation methods, clearly highlighting the exceptional capability of EITNet in addressing complex multi-view basketball action recognition tasks. Specifically, EITNet achieved an accuracy of 90.2\% in the Cross-Subject evaluation and 92.1\% in the Cross-View evaluation, both of which surpass the performance of other recently proposed advanced models. These results demonstrate that EITNet has a stronger generalization ability and robustness in handling variations across different subjects and multi-view scenarios. In contrast, other models showed relatively weaker performance in both evaluation settings. Notably, the U-STN and PI3D models, although demonstrating certain advantages in specific evaluation scenarios, still lag behind EITNet in overall accuracy. This discrepancy may be attributed to EITNet's more effective utilization of multimodal data fusion and its advanced deep learning architecture, enabling it to maintain higher recognition precision in highly complex and dynamic environments. Overall, the EITNet model not only demonstrates theoretical innovation but also shows exceptional performance in practical applications, validating its effectiveness for complex action recognition tasks. These findings further underscore that integrating multi-view video analysis with deep learning technologies can significantly enhance the overall performance of action recognition models.

\begin{table}[h!]
\centering
\caption{Comparison of EITNet with Other Models on Cross-Subject and Cross-View Evaluations}
\resizebox{\columnwidth}{!}{
\begin{tabular}{lcc}
\toprule
Model & \multicolumn{2}{c}{Accuracy (\%)} \\
\cmidrule(lr){2-3}
 & Cross-Subject Evaluation & Cross-View Evaluation \\
\midrule
EITNet (Ours) & 90.2 & 92.1 \\
U-STN~\cite{yang2022u} & 84.6 & 82.9 \\
PI3D~\cite{wu2021pose} & 85.2 & 83.4 \\
MADT-GCN~\cite{xia2024skeleton} & 86.8 & 84.7 \\
SF-GCN~\cite{sun2021multi} & 88.3 & 86.1 \\
TOQ-Nets~\cite{mao2021temporal} & 89.1 & 87.3 \\
\bottomrule
\end{tabular}
}
\label{cross}
\end{table}

\subsubsection{Model Complexity Comparison}
\begin{table}[h]
\centering
\caption{Comparison of Model Complexity in Terms of Parameters and FLOPs}
\resizebox{\columnwidth}{!}{
\begin{tabular}{lcc}
\toprule
Model       & Parameters (Millions) & FLOPs (GFLOPs) \\
\midrule
Pavllo et al.~\cite{pavllo20193d}         & 16.95                          & 33.87                   \\
Chen et al.~\cite{chen2021anatomy}          & 31.8          & 1.78                    \\
P-STMO~\cite{shan2022p}               & 6.2                            & 1.4                     \\
GTFormer~\cite{huo2023gtformer}              & 19.04                          & 2.31                    \\
MHFormer~\cite{li2022mhformer}              & 25.2                           & 1.38                    \\
EITNet                & 12.5                           & 1.2                     \\
\bottomrule
\end{tabular}
}
\label{complexity}
\end{table}

As shown in Table \ref{complexity}, EITNet demonstrates a clear advantage in terms of both parameter count and computational complexity. Specifically, EITNet has 12.5 million parameters, significantly fewer than other models such as Chen et al. with 31.8 million and MHFormer with 25.2 million. This indicates that EITNet maintains a simpler model structure. In terms of FLOPs, EITNet has a computational cost of 1.2 GFLOPs, the lowest among all models, and particularly much lower than Pavllo et al., which requires 33.87 GFLOPs. This low computational cost means that EITNet can operate at a lower computational burden while still maintaining high performance. These results demonstrate that EITNet not only achieves greater computational efficiency in model design but also performs effectively in handling complex tasks. Its reduced parameter count and FLOPs make it highly suitable for resource-constrained environments, particularly in real-time multi-view basketball action recognition tasks, where faster response times and higher operational efficiency are critical.

\subsubsection{Accuracy and Loss Evaluation}

\begin{figure}[h]
	\centering
	\includegraphics[width=0.5\textwidth, height=0.18\textheight]{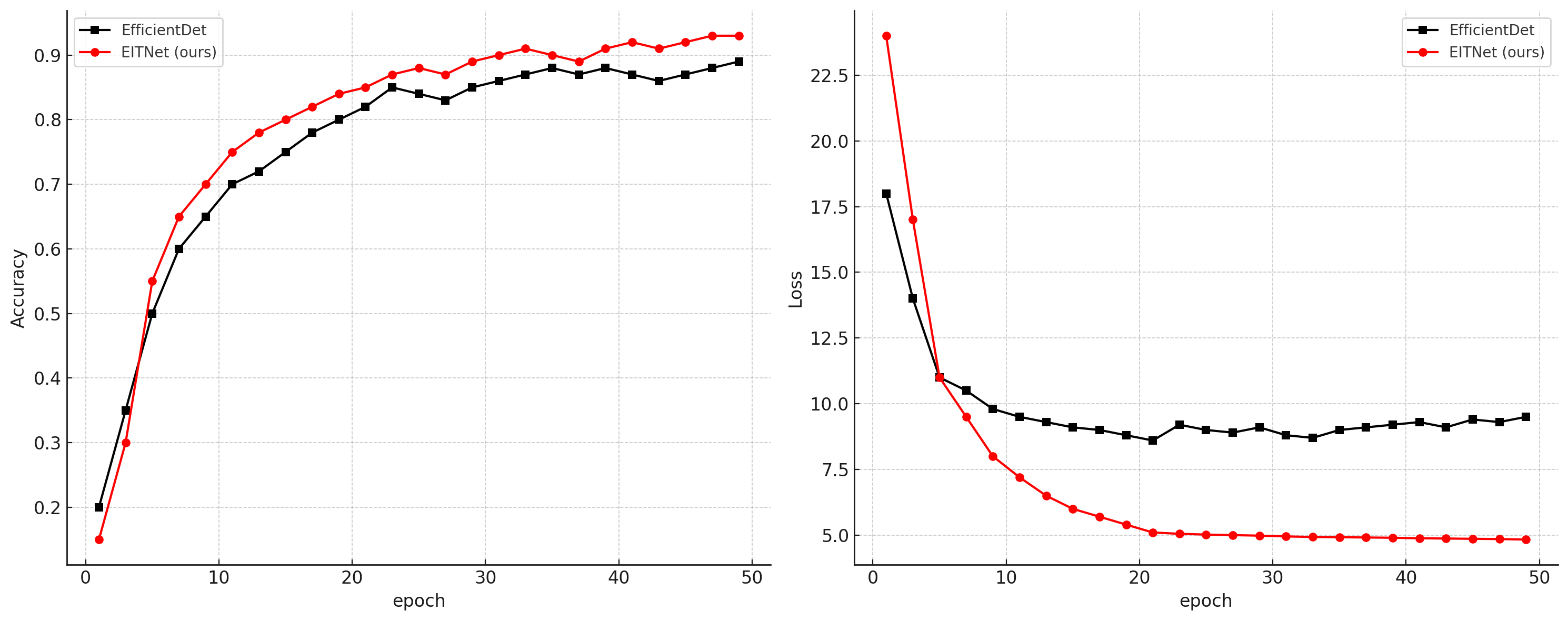}
	\caption{Accuracy and loss compariso over 50 epochs.}
	\label{loss}
\end{figure}

The evaluation of EITNet and EfficientDet models across 50 epochs, as shown in Figure \ref{loss}, demonstrates EITNet’s superior learning and generalization capability. The accuracy curve shows that EITNet reaches 0.92, approximately 5\% higher than EfficientDet, which stabilizes at around 0.87. Notably, EITNet maintains a consistent upward trend after the 10th epoch, while EfficientDet's accuracy fluctuates, indicating EITNet's more effective pattern recognition and adaptability during training.
Regarding loss, EITNet shows a more efficient reduction from 22.5 to below 5.0, stabilizing after 20 epochs. In contrast, EfficientDet’s loss decreases more slowly and stabilizes at a higher value of around 9.0, suggesting that EITNet achieves faster convergence and a better overall fit with fewer errors. These results confirm that EITNet’s architecture significantly improves accuracy and loss reduction, making it more effective for reliable basketball action recognition in real-world scenarios.

\subsubsection{Ablation Study}

\begin{table*}[h!]
\centering
\caption{Ablation Study Results: MPJPE Comparison on CMU Panoptic Studio and NPU RGB+D Datasets}
\begin{tabular}{lccccc}
\toprule
Configuration & EfficientDet & I3D & TimeSformer & MPJPE (mm) & MPJPE (mm) \\
 &  &  &  & CMU Panoptic Studio & NPU RGB+D \\
\midrule
EITNet (Full Model) & \checkmark & \checkmark & \checkmark & 45.3 & 38.6 \\
Without EfficientDet & \texttimes & \checkmark & \checkmark & 49.2 & 42.3 \\
Without I3D & \checkmark & \texttimes & \checkmark & 47.6 & 40.9 \\
Without TimeSformer & \checkmark & \checkmark & \texttimes & 46.8 & 39.7 \\
\bottomrule
\end{tabular}
\label{ablation}
\end{table*}

Table \ref{ablation} presents the performance of the EITNet model under various configurations, highlighting the impact of each component on the overall model performance. The ablation study results indicate that the complete EITNet model achieved MPJPE values of 45.3 mm and 38.6 mm on the CMU Panoptic Studio and NPU RGB+D datasets, respectively, outperforming the configurations where any single component was removed. Specifically, removing the EfficientDet module resulted in MPJPE values increasing to 49.2 mm and 42.3 mm on the respective datasets, demonstrating the crucial role this module plays in object detection and precise localization of player actions. Similarly, the absence of the I3D module led to an increase in MPJPE values to 47.6 mm and 40.9 mm, underscoring the importance of I3D in extracting spatiotemporal features. Removing the TimeSformer module increased MPJPE values to 46.8 mm and 39.7 mm, which confirms the importance of temporal sequence analysis for maintaining high accuracy in action classification. These results clearly illustrate that each component of EITNet is essential for effectively analyzing multi-view video data and recognizing complex action patterns, thereby enhancing the precision and reliability of movement evaluation.

\subsubsection{Action Recognition Results and Analysis}

\begin{figure*}[h]
	\centering
	\includegraphics[width=1\textwidth, height=0.4\textheight]{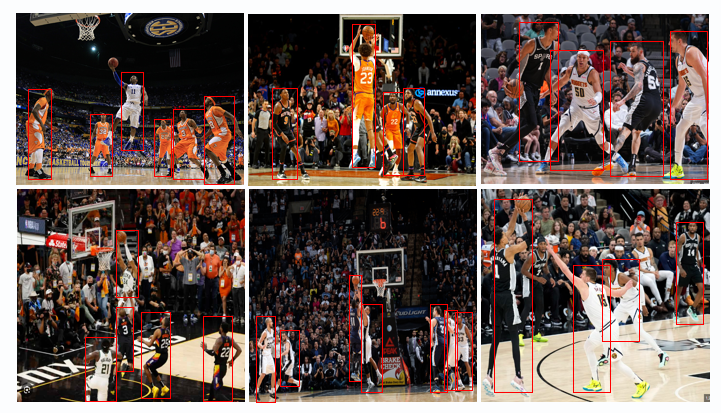}
	\caption{Examples of basketball action recognition using the EITNet model, highlighting accurate detection of various player actions across different game scenarios.}
	\label{visual}
\end{figure*}

Figure \ref{visual} demonstrate the effectiveness of the EITNet model in accurately detecting and recognizing various basketball player actions. The red bounding boxes highlight the detected players and their respective actions, such as shooting, dribbling, and passing, under different game scenarios. The consistent and accurate detection across diverse and dynamic scenes, even with multiple players and varying postures, illustrates the robustness of the EITNet model in handling complex real-world basketball environments. This performance underscores the model's capability to generalize well across different contexts, making it a reliable tool for action recognition in sports analytics.

\section{Conclusion}

In this study, we introduced the EITNet model, a sophisticated deep learning architecture designed for effective basketball player action recognition and analysis. The model integrates advanced components such as EfficientDet for object detection, I3D for spatiotemporal feature extraction, and TimeSformer for temporal analysis, along with IoT technology to enable real-time data acquisition and processing. Extensive experiments demonstrated that EITNet significantly outperformed the baseline EfficientDet model, achieving an approximately 5\% higher accuracy and reducing loss by over 50\% throughout multiple training epochs. The inclusion of IoT devices facilitated seamless data collection and transmission, thereby enhancing the system's operational capability in real-world scenarios. The model was rigorously tested in dynamic basketball environments, where it effectively detected and classified a wide range of player actions, confirming its robustness and adaptability to complex conditions.

However, the EITNet model does have certain limitations. The integration of IoT and advanced deep learning components results in a higher computational load, which poses challenges for deployment on devices with limited processing power, such as mobile or edge devices. Additionally, this increased computational demand leads to longer training times, which may not be ideal for applications requiring rapid updates or real-time analysis. The model may also encounter difficulties in scenarios involving extreme occlusion or complex player interactions, where the IoT sensors and cameras might not fully capture the nuances of the actions, potentially leading to misclassification or reduced accuracy, especially in situations where players are closely interacting or obscured from view. Furthermore, when applied to other real-world applications, such as different sports or activities, the model might struggle to adapt to varying movement patterns, diverse environmental conditions, and differences in the quality of available data, which could limit its effectiveness and generalization across various domains.

To address these limitations, future research should focus on optimizing the EITNet model to reduce its computational complexity, making it more efficient for real-time applications and deployment across various devices. This could involve exploring model compression techniques, such as pruning and quantization, or developing more efficient neural architectures that maintain high performance with lower resource requirements. Furthermore, enhancing the model's robustness against occlusion and complex interactions could be achieved by upgrading the IoT system with more sophisticated sensors or incorporating additional data sources, such as biometric information or contextual game data. These advancements would not only strengthen the EITNet model but also contribute to the broader field of sports analytics, creating new opportunities for automated sports analysis and player performance improvement. Ultimately, the combination of IoT and advanced deep learning techniques in the EITNet model represents a significant step forward in the development of intelligent sports analysis systems.

\section*{Author Contributions}
Jingyu Liu was responsible for the conceptualization and overall design of the study, as well as leading the manuscript writing. Xinyu Liu conducted the experiments and performed data analysis. Mingzhe Qu contributed to the development and implementation of the IoT technology and the multiview video analysis framework, and coordinated the research activities. Tianyi Lyu was involved in the integration of the system components and optimization of the methodology. All authors contributed to the final version of the manuscript.

\section*{Data availability}
The data and materials used in this study are not currently available for public access. Interested parties may request access to the data by contacting the corresponding author.

\section*{conflicts of interest}
The authors declare that the research was conducted in the absence of any commercial or financial relationships that could be construed as a potential conflict of interest.

\bibliographystyle{cas-model2-names}
\bibliography{cas-refs}


\end{document}